\documentclass[conference,10pt]{IEEEtran}
\IEEEoverridecommandlockouts
\usepackage{cite}
\usepackage{amsmath,amssymb,amsfonts}
\IfFileExists{algorithmic.sty}{\usepackage{algorithmic}}{}
\usepackage{graphicx}
\usepackage{textcomp}
\usepackage[table]{xcolor}
\usepackage{booktabs}
\usepackage{multirow}
\usepackage{subcaption}

\begin{document}

\title{Contribution-Aware Bandwidth Allocation for Multimodal Split Learning}

\author{

\IEEEauthorblockN{Anonymous}
}

\author{
    \IEEEauthorblockN{Iason Ofeidis, 
    Leandros Tassiulas}
    \IEEEauthorblockA{
        Department of Electrical and Computer Engineering, Yale University, New Haven, CT, USA
    }
}

\maketitle

\begin{abstract}
    Multimodal models are increasingly the default option for perception at the network
    edge, yet they are trained almost entirely in the datacenter, because a client
    holding several sensor streams cannot host an encoder per modality.
    Split Learning makes such training feasible by keeping only the first layers on
    the device, at the cost of an uplink that must carry smashed activations for
    every modality at every step.
    Existing compression schemes give each modality the same keep-ratio, so the
    shared budget is divided in proportion to smashed-activation dimension, a
    quantity unrelated to how much each modality contributes to the fused
    prediction.
    We make that division an explicit decision and call it \emph{inter-modality allocation}: under a fixed uplink budget, every policy transmits the same expected payload and differs only in how that payload is split across modalities. 
    Our allocator, \textnormal{ModalShare}, sets each modality's keep-ratio from a Shapley contribution score that the server computes over coalitions of activations it has already received. 
    Measuring this score adds no uplink traffic and no client-side computation, and needs no prior knowledge of which stream is which.
    \textnormal{ModalShare} improves accuracy over equal
    keep-ratios by \(15.4\) and \(12.4\) percentage points on CREMA-D and MVSA at
    matched payload in $5\times$ compression, with strong performance across three 
    compressors, three datasets, and four budgets.
    We show that existing compressors underperform in multimodal settings, with \textnormal{ModalShare} recovering what gains are left behind.
\end{abstract}


\section{Introduction}
\label{sec:intro}

Artificial intelligence is increasingly multimodal.
Frontier models now match or exceed human baselines on multimodal reasoning,
while the sensors that feed such models (e.g. cameras, microphones, accelerometers) 
are already standard on the devices people carry.
Emotion recognition from speech and facial expression, sentiment analysis over
paired image and text, and activity recognition from inertial streams all
depend on combining modalities rather than choosing among them.
Yet the models that do this combining are trained and served almost entirely in
the cloud, while the data that would train them best originates at the
edge.
Closing that gap means running multimodal learning where the sensors are, and
that is precisely where the deployment assumptions break down: an encoder per
modality multiplies memory and compute against a budget that was already tight
before the second stream arrived.

Federated Learning~\cite{mcmahan2017communication} keeps raw data on the device, but leaves this problem
unaddressed, since the full model must still fit locally.
Split Learning (SL)~\cite{vepakomma2018split} removes the constraint by partitioning the model at a cut
layer: the client runs only the first few layers and
transmits the resulting smashed activations to a server that completes the
forward and backward passes.
The client-side share of the model can then be made as small as the deployment
demands, the scheme composes with federated aggregation, and it appears
repeatedly in 6G edge architectures for exactly this
reason~\cite{lin2024split,hafi2024split}.

This gain comes from trading on-device compute for bandwidth, and the cost concentrates on the uplink, the more constrained of the two directions.
The cut-layer uplink carries per-batch activations for the entire training run,
and their size is the dominant cost of the
system~\cite{lin2024split,oh2025communication}.
Uplink is also the scarce direction in practice: most operators now measure
uplink traffic growing faster than downlink, and AI services are expected to
drive uplink volumes to roughly three times their 2025 level by
2031~\cite{ericsson2026}.
A substantial literature therefore compresses what crosses the cut: Top-\(K\)
sparsification~\cite{yuan2020federated}, randomized selection that reduces
index overhead~\cite{zheng2023reducing}, progressive pruning of the
representation over training~\cite{mudvari2024adaptive}, channel-wise rate
adaptation~\cite{lin2026sl}, and adaptive feature-wise dropout at the
cut~\cite{oh2025communication}.
Each of these decides how a \emph{single} tensor (e.g. activations) is represented within its
budget, which is the only decision available when a node holds one tensor.
Applied to a multimodal client, they are modality-agnostic by construction: one
policy, and the same keep-ratio handed to every stream.

\begin{figure}[t]
  \centering    
  \includegraphics[width=\linewidth]{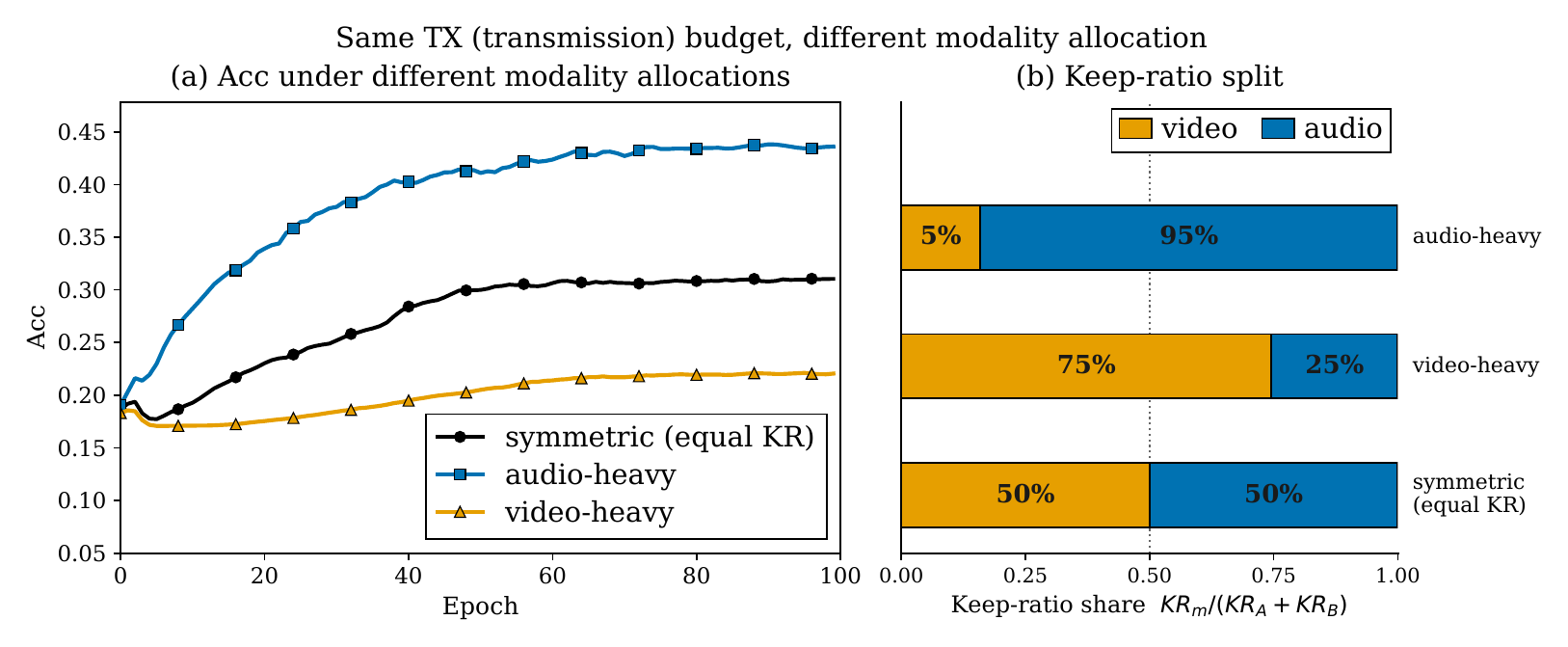}
  \caption{Fixed-budget keep-ratio allocations across modalities
  (CREMA-D, \(5\times\) compression). All three arms transmit the
  same total payload under the same compressor and differ only in how it is
  divided (b); the resulting accuracy differs by a wide margin (a). Allocation,
  not only budget and compression, determines what a fixed uplink buys.}
  \label{fig:combined}
\end{figure}

A multimodal client, however, holds several tensors that share one uplink, and
that raises a second decision the compression literature does not expose: how
the shared budget is divided among them.
Equal keep-ratios divide the transmitted payload in proportion to the cut
dimensions, so a modality whose smashed activations are four times larger
silently receives four times the bandwidth: equal compression severity is not
equal bandwidth.
For example, on MVSA-Single, an image-text sentiment dataset, the image cut is roughly four 
times the text cut, so image takes $80\%$ of the budget before anything 
about the task is measured.
Additionally, no evidence suggests that a dimension-based division matches how the modalities contribute; in fact, modalities are weighted unequally in fused models and degrade differently under constraint~\cite{xu2025contribution,wei2025improving,peng2022balanced}.
Under a fixed budget, floats spent on a stream the fused prediction barely uses
are floats taken from one it depends on.
Figure~\ref{fig:combined} indicates that the difference is not marginal
even when dimension skew is absent:
on CREMA-D, an audio-video emotion recognition task whose two cuts are
nearly equal in dimension, holding the transmitted payload and the
compressor fixed and varying only how the budget is split, an audio-heavy
allocation converges well above the equal-keep-ratio default while a
video-heavy one falls below it.
Only the allocation differs.

This motivates the question we study: under a fixed total uplink budget, does
allocating bandwidth asymmetrically across modalities outperform uniform allocation,
and under what conditions?
We make the decision explicit as \emph{inter-modality allocation} and design a
policy for it.
Our allocator, \textit{ModalShare}, estimates each modality's marginal contribution with
a Shapley probe evaluated at the server over coalitions of smashed activations
the server already holds, and maps those scores to per-modality keep-ratios
that preserve the total transmitted payload.
Because the coalitions are formed by masking activations that have already
crossed the cut, the measurement consumes no uplink, has no client-side 
computation overhead, 
and requires no auxiliary
channel; its cost is only server-side compute.
Allocation sits above the compressor rather than replacing it: ModalShare
determines how much budget each stream receives, while the existing compressor
still determines which coordinates within that stream survive.
We find the resulting gains to be substantial: largest at
moderate compression on cuts where contribution is skewed, diminishing as the
budget approaches the trivial floor, and near zero where the modalities
genuinely contribute equally, which is the behaviour a measurement-driven
policy should exhibit.

\noindent The main contributions of this work are:
\begin{itemize}
\item We show that modality-agnostic compression is not neutral: equal
      keep-ratios allocate transmitted payload in proportion to cut dimension,
      a quantity unrelated to a modality's role in the fused prediction.
      To our knowledge, this is the first work to treat the division of a
      shared cut budget across modalities as a decision in its own right, which
      we formulate as \emph{inter-modality allocation}: an iso-budget problem
      in which every policy transmits the same expected payload and differs
      only in its split.

\item We propose \emph{ModalShare}, which sets per-modality keep-ratios from a
      fusion-level contribution score measured at the server over coalitions of
      already-received smashed activations, adding no uplink traffic and
      assuming no modality-identity prior.
      Under SplitFC at \(\beta{=}0.2\) it improves accuracy over equal
      keep-ratios by \(15.4\)\,pp on CREMA-D and \(12.4\)\,pp on MVSA at
      matched payload, with the 
      advantage persisting across three compressors, three datasets, and four 
      budgets; existing SL compressors leave these gains
      unclaimed in multimodal settings, and allocation recovers them.

\item We attribute the gain to contribution rather than to dimension skew.
      On MVSA the contribution component exceeds the skew component that any
      fixed rule could recover, and on CREMA-D, whose cut dimensions are
      balanced, it accounts for the whole effect; where the measured gap is
      small, as on UCI~HAR, allocation stays near-uniform and the gain
      vanishes.

\item We show cut-local proxies do not substitute for a fusion-level score:
      activation and gradient norms cannot observe how modalities combine, and
      standalone utility measures a modality in isolation rather than its
      marginal value, so none recovers the gain.
\end{itemize}

\section{Related Work}
\label{sec:related}

\noindent \textbf{Split Learning \& Compression.}
SL partitions a model so that clients transmit intermediate
activations to a server that completes the forward and backward
passes~\cite{vepakomma2018split}, and the size of those intermediates is the
dominant cost~\cite{oh2025communication,lin2024split}.
Communication-efficient variants of SL compress the transmitted tensor: Top-\(K\)
sparsification~\cite{yuan2020federated}, randomized variants that reduce
selection and index cost~\cite{zheng2023reducing}, progressive pruning of the
representation over training~\cite{mudvari2024adaptive}, adaptive
feature-wise dropout at the cut~\cite{oh2025communication}, channel-wise
compression that adapts the rate across channels of a single
tensor~\cite{lin2026sl}, batch-wise packing of multiple smashed vectors into one~\cite{hsieh2022c3}, 
and learned autoencoder bottlenecks at the
cut~\cite{meuwissen2026autoencoder, shao2020bottlenet++}.
Each of these decides how a single tensor is represented: a sparsity level, a
schedule, a channel-wise rate or a learned bottleneck.
Multimodality adds a second: when several streams share one uplink, the budget
must also be divided among them.
That division is the subject of this work.

\noindent \textbf{Asymmetric Multimodal Learning.}
Modality imbalance in centralized training has received increasing attention in recent years.
Gradient-balancing methods such as on-the-fly gradient modulation
(OGM)~\cite{peng2022balanced} equalize optimization effort across modalities.
Wei et al.~\cite{wei2025improving} argue that balanced optimization is
not optimal and instead align each modality's optimization dependency with the
inverse of its variance ratio.
Xu et al.~\cite{xu2025contribution} guide asymmetry by an explicit contribution
score, using it both to accelerate gradients and to set a per-modality
information-bottleneck strength; greedy modality
selection~\cite{cheng2022greedy} makes the coarser decision of which modalities
to admit at all.
Modality-balanced quantization~\cite{li2025mbq} makes a related observation at
the level of numerical precision, calibrating vision and language tokens
separately because a single quantization objective serves them unequally.
These works reallocate optimization effort, representation capacity, or
numerical precision across modalities.
We reallocate a communication budget at the split-learning cut, where the
constraint is a transmitted payload rather than a regularizer.

\noindent \textbf{Contribution Scoring.}
Shapley-style attribution quantifies how much each modality drives a prediction, and has been used to audit vision-language transformers~\cite{parcalabescu2023mm}, to separate unimodal from cross-modal contributions~\cite{hu2022shape}, and to report diagnostic weight in clinical models~\cite{soenksen2022integrated}.
Closest in estimator, Wei et al.~\cite{wei2024enhancing} compute Shapley-based contributions per sample and resample data for the weaker modality.
Contribution is consistently found to be dataset- and task-dependent, which is what makes a fixed budget split unsatisfactory and an online estimate useful.
The question also arises in federated learning, where contribution estimates have been applied to communication~\cite{yuan2026communication} and to modality imbalance across clients~\cite{amini2025distributed}; there the network carries aggregated model updates rather than per-batch activations at a fixed cut, so the allocation decision we study does not arise.
We evaluate the same coalitions during training and spend the resulting scores on transmitted payload: the estimator is shared, the actuator is the uplink budget.
To our knowledge, this is the first use of a Shapley-based contribution score to allocate communication resources across modalities in multimodal learning.

\section{Method}
\label{sec:method}

We consider multimodal SL between one multimodal client and one
server: the client holds the early modality encoders, the server holds the
remaining backbone, fusion, and classifier.
Because attribution and allocation act only on that client's cut and uplink,
the same policy applies independently on every client--server link in a
multi-client deployment.
We formulate sharing the client--server cut as an iso-budget allocation problem
and give a contribution-aware policy for setting per-modality budgets.

\begin{table}[t]
  \centering
  \caption{Notation.}
  \label{tab:notation}
  \setlength{\tabcolsep}{4pt}
  \renewcommand{\arraystretch}{1.05}
  \begin{tabular}{@{}cl@{}}
  \toprule
  Symbol & Meaning \\
  \midrule
  \(M\) & number of modalities \\
  \(D_m\) & flattened cut dimension of modality \(m\) \\
  \(z_m\) & smashed activations of modality \(m\) \\
  \(r_m\) & keep-ratio of modality \(m\) (the allocated quantity) \\
  \(\beta\) & uplink budget as a fraction of full cut payload,
              \(\mathrm{TX}/\sum_k D_k\) \\
  \(q_m\) & payload share of modality \(m\), \(D_m r_m/\mathrm{TX}\) \\
  \(\mathcal{C}\) & intra-modality compressor (SplitFC primary) \\
  \(u(S)\) & coalition utility (server true-label log-prob with only
             modalities \(S\)) \\
  \(\phi_m\) & Shapley contribution score of modality \(m\) \\
  \(s_m\) & EMA-smoothed contribution share of modality \(m\) \\
  \(\pi_m\) & normalized allocation weight from \(s_m\) \\
  \(\tau\) & allocation temperature \\
  \(r_{\min}\) & keep-ratio floor \\
  \(g\) & contribution confidence gap \\
  \(\gamma\) & freeze confidence threshold \\
  \(K\) & consecutive confident epochs required to freeze \\
  \(e_{\min}\) & earliest epoch at which freezing is allowed \\
  \bottomrule
  \end{tabular}
  \end{table}

\subsection{Multimodal Split Learning}
\label{sec:setup}

Let \(\mathcal{D}=\{(x_i^{(1)},\ldots,x_i^{(M)},y_i)\}_{i=1}^{N}\) be a training
set with modality index set \(\mathcal{M}=\{1,\ldots,M\}\).
A single client hosts \(M\) modality encoders; for modality \(m\), encoder
\(g_m\) with parameters \(\theta_m\) maps its input to intermediate activations
(smashed activations) at a fixed cut,
\begin{equation}
  z_m = g_m(x^{(m)};\theta_m) \in \mathbb{R}^{D_m},
  \label{eq:smashed}
\end{equation}
where \(D_m\) is the flattened smashed-activation dimension.
The server model \(h\) with parameters \(\theta_s\) receives the smashed
activations, completes any remaining modality-specific layers, fuses the
modalities, and predicts
\begin{equation}
  \hat y = h(\{z_m\}_{m\in\mathcal{M}};\theta_s).
  \label{eq:server}
\end{equation}
For each mini-batch the client uploads \(\{z_m\}\); the server evaluates the
task loss, backpropagates through \(h\), and returns the cut gradients
\(\nabla_{z_m}\ell\).
All communication crosses this cut.

\subsection{Compression of smashed activations}
\label{sec:compression}

Let \(\mathcal{C}\) denote a modality-level compressor with budget parameter
\(r_m\in(0,1]\),
\begin{equation}
  \hat z_m = \mathcal{C}(z_m,r_m),
  \label{eq:compressor}
\end{equation}
where the keep-ratio \(r_m\) is the expected payload of \(\hat z_m\) as a
fraction of \(D_m\).
This interface separates two decisions: the compressor determines \emph{how} to
represent modality \(m\) within its budget, and the allocator determines
\emph{how much} budget that modality receives.
Our allocation rule is therefore independent of the internal selection rule
used by \(\mathcal{C}\).

Budget is accounted in transmitted smashed-activation floats,
\begin{equation}
  \mathrm{TX} = \sum_{m\in\mathcal{M}} D_m r_m,
  \label{eq:tx}
\end{equation}
per sample and per direction.
The compressed representation also defines the units the server can
differentiate with respect to, so the return path carries gradients in that
same representation: coordinates the client did not send receive no gradient,
and total cut traffic is \(2\cdot\mathrm{TX}\).
Two allocations at equal \(\mathrm{TX}\) therefore exchange the same expected
payload in both directions, and whatever side information \(\mathcal{C}\)
carries with it (e.g., masks and indices for a sparsifier, scales for a
quantizer), is matched along with it.

\begin{figure*}[t]
  \centering
  \includegraphics[width=\textwidth]{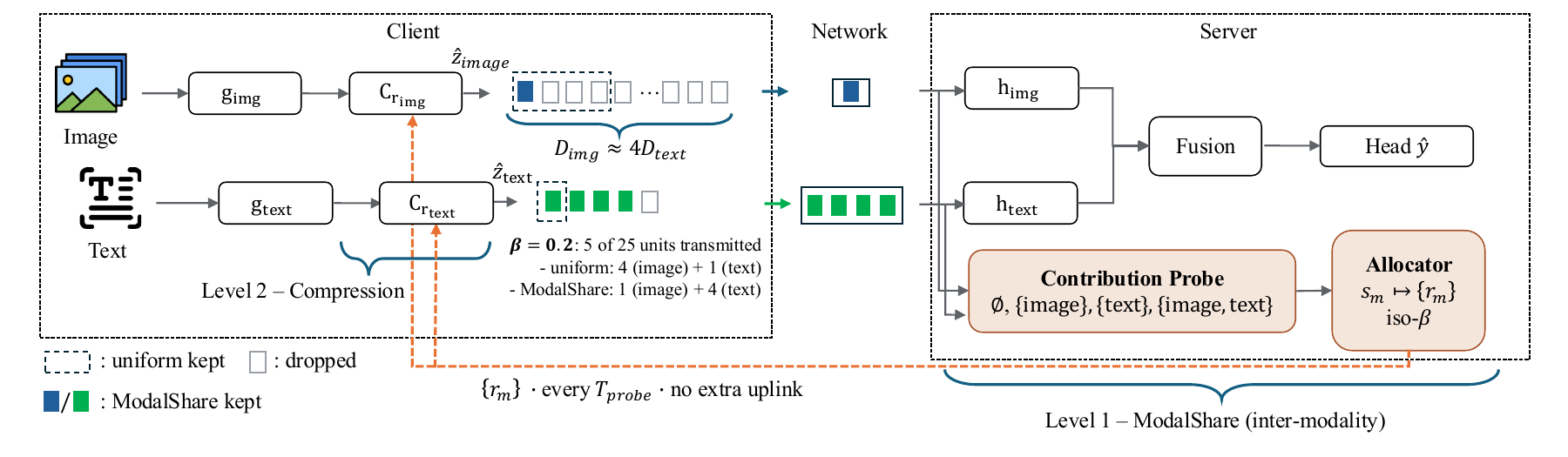}
  \caption{\textbf{ModalShare allocates a fixed cut budget across modalities}.
  Prior compression operates within a modality (Level 2); ModalShare divides the
  shared payload between them (Level 1) using a server-side Shapley probe on
  already-received activations, so no extra uplink is consumed. Strip lengths
  follow the MVSA cut dimensions; cell counts are illustrative at \(\beta=0.2\),
  where both policies transmit 5 of 25 units and differ only in the split.}
  \label{fig:method}
  \end{figure*}

\subsection{Inter-modality allocation}
\label{sec:simplex}

Existing compression of smashed activations applies one rule to concatenated
smashed activations or assigns the same keep-ratio to every modality.
Neither asks how a shared budget should be divided according to the modalities'
roles in the fused prediction.
We expose that decision as inter-modality allocation.

We define the normalized communication budget as:
\begin{equation}
  \beta = \frac{\mathrm{TX}}{\sum_m D_m}.
  \label{eq:txfrac}
\end{equation}
All policies compared at budget \(\beta\) satisfy the iso-budget constraint
\begin{equation}
  \sum_m D_m r_m = \beta\sum_m D_m,
  \label{eq:isobudget}
\end{equation}
so they exchange the same expected payload and differ only in its division.
The payload share of modality \(m\) is
\begin{equation}
  q_m = \frac{D_m r_m}{\mathrm{TX}},
  \qquad \sum_m q_m=1,
  \label{eq:q}
\end{equation}
and \(\mathbf{q}\) lies on the probability simplex, giving a dimension-aware
description of allocation along an iso-budget surface.

Equal keep-ratios, \(r_m=\beta\), give
\begin{equation}
  q_m^{\mathrm{uni}} = \frac{D_m}{\sum_k D_k}.
  \label{eq:q-unif}
\end{equation}
Equal compression severity therefore does not imply equal transmitted floats:
a modality with four times the smashed-activation dimension receives four times
the payload.
Our objective is to choose \(\mathbf{q}\) from fusion-level contribution while
preserving~\eqref{eq:isobudget}.

\subsection{ModalShare: Contribution-aware allocation}
\label{sec:ours}

Here, we introduce \emph{ModalShare}, a contribution-aware allocator that divides
a fixed uplink budget into
per-modality keep-ratios based on a server-side contribution score, leaving the
intra-modality compressor unchanged
(Figure~\ref{fig:method}).

\noindent \textbf{Server-side contribution probe.}
We define the coalition utility \(u(S)\) as the mean true-label log-probability
produced by the server when only modalities \(S\subseteq\mathcal{M}\) are
present, with absent modalities replaced by zeros, a masking convention used also
in modality valuation~\cite{wei2024enhancing}.
Coalition marginals are aggregated by the Shapley value~\cite{shapley1953value},
\begin{equation}
  \phi_m =
  \sum_{S\subseteq\mathcal{M}\setminus\{m\}}
  \frac{|S|!\,(M-|S|-1)!}{M!}
  \left[u(S\cup\{m\})-u(S)\right],
  \label{eq:shapley}
\end{equation}
evaluated exactly with \(2^M\) coalitions, or by sampling coalition orderings
for large \(M\).
For example, when \(M=2\) the four coalitions are
\(\emptyset,\{1\},\{2\},\{1,2\}\).

Because batch-level utilities are noisy, negative contributions are floored,
renormalized to a share, and then smoothed,
\begin{equation}
  \tilde\phi_m^{(t)}
  =
  \frac{[\phi_m^{(t)}]_+ }{\sum_k [\phi_k^{(t)}]_+ + \varepsilon},
  \qquad
  s_m^{(t)} = \alpha\tilde\phi_m^{(t)} + (1-\alpha)s_m^{(t-1)},
  \label{eq:ema}
\end{equation}
with the probe evaluated every \(T_{\mathrm{probe}}\) batches and cached scores
used in between.

\noindent \textbf{From contribution to keep-ratios.}
Given scores \(\{s_m\}\), temperature-controlled weights
\begin{equation}
  w_m=s_m^{1/\tau},
  \qquad
  \pi_m=\frac{w_m}{\sum_k w_k},
  \label{eq:allocation-weights}
\end{equation}
convert contribution into a share of the residual payload after a keep-ratio
floor \(r_{\min}\) reserves \(D_m r_{\min}\) units per modality.
With \(\mathrm{TX}_{\mathrm{sur}}=\mathrm{TX}-r_{\min}\sum_k D_k>0\),
\begin{equation}
  c_m=D_m r_{\min}+\pi_m \mathrm{TX}_{\mathrm{sur}},
  \qquad
  r_m=\frac{c_m}{D_m},
  \label{eq:allocation-map}
\end{equation}
followed by projection to \(r_m\in[r_{\min},1]\) that preserves the total
budget~\eqref{eq:isobudget}.
Smaller \(\tau\) sharpens~\eqref{eq:allocation-weights}.
The empty-coalition utility sets the scale of \(\{\phi_m\}\): for \(M{=}2\),
\(\phi_1-\phi_2=u(\{1\})-u(\{2\})\) is independent of \(u(\emptyset)\), while
efficiency gives \(\sum_m\phi_m=u(\mathcal{M})-u(\emptyset)\), so the baseline
controls how sharply the weights separate the modalities.
When the keep-ratio floor vanishes, a decisive contribution gap can place the
allocation at a simplex corner rather than a smooth interior reweighting.
Hard modality selection is the limiting case $\tau \rightarrow 0$ of this map; the floor $r_{min}$ and the temperature $\tau$ interpolate between it and uniform allocation.

\noindent \textbf{Confidence and freeze.}
Live allocation follows~\eqref{eq:allocation-map} without gating keep-ratios on
the contribution gap: while unfrozen, ModalShare remains adaptive.
Confidence is measured separately by the normalized gap between the leading
and runner-up scores,
\begin{equation}
  g
  =
  \frac{s_{(1)}-s_{(2)}}{\sum_k s_k + \varepsilon},
  \label{eq:confidence-gap}
\end{equation}
where \(s_{(1)}\ge s_{(2)}\) are the top two values in \(\{s_m\}\).
For \(M{=}2\) this reduces to
\(g=\lvert s_1-s_2\rvert/(s_1+s_2+\varepsilon)\).
After epoch \(e_{\min}\), if \(g\ge\gamma\) favors the same winning modality for
\(K\) consecutive epochs, ModalShare freezes the current keep-ratios
\(\{r_m\}\), preserves~\eqref{eq:isobudget}, and discontinues probing.
A drop of \(g\) below \(\gamma\), or a change of winner, resets the streak.
The threshold \(\gamma\) therefore controls only when to stop adapting, not the
online map itself.

\subsection{Training procedure and communication cost}
\label{sec:train}

\begin{figure}[t]
\centering
\begin{minipage}{0.92\linewidth}
\small
\noindent\textbf{Algorithm 1} ModalShare iso-budget allocation.
\\[0.4em]
\hrule height 0.6pt
\vspace{0.35em}
\begin{enumerate}\setlength{\itemsep}{0.15em}
  \item Compute smashed activations
        \(z_m\gets g_m(x^{(m)})\), \(m\in\mathcal{M}\).
  \item Every \(T_{\mathrm{probe}}\) batches while unfrozen: evaluate server
        coalitions, update \(\{s_m\}\), and map to \(\{r_m\}\) under
        \eqref{eq:allocation-map}.
  \item If the freeze criterion on \(g\) is met, hold \(\{r_m\}\) fixed and
        skip further probes; otherwise continue adapting.
  \item The client compresses and uploads
        \(\hat z_m\gets\mathcal{C}(z_m,r_m)\).
  \item Complete server forward/backward, return cut gradients, and update
        \(\{\theta_m\}\) and \(\theta_s\).
\end{enumerate}
\vspace{0.2em}
\hrule height 0.6pt
\end{minipage}
\end{figure}

Algorithm~1 embeds the probe, map, and freeze rule in the training loop.
During early training the uplink budget is ramped from full keep-ratios
(\(r_m{=}1\)) to the target fraction \(\beta\); schedules appear in
Section~\ref{sec:experimental_setup}.
While unfrozen, ModalShare refreshes \(\{r_m\}\) via~\eqref{eq:allocation-map}
on the probe schedule of Section~\ref{sec:ours}, including during the
compression ramp.
At fixed \(\beta\), compared policies share the same expected \(\mathrm{TX}\) and
differ only in \(\mathbf{q}\).

The probe adds no client--server messages: coalitions are formed by masking
smashed activations already held at the server, so attribution consumes no
part of~\eqref{eq:tx} and needs no auxiliary channel.
Its cost is \(2^M\) forward evaluations of the server submodel (four for
\(M{=}2\)) every \(T_{\mathrm{probe}}\) batches while allocation remains unfrozen.
Freezing removes that overhead for the remainder of training once the
confidence criterion is met.
Section~\ref{sec:results-freeze} compares the frozen policy to continued
online allocation.

\section{Experimental Setup}
\label{sec:experimental_setup}

Our evaluation tests an allocation hypothesis rather than a new feature
compressor.
At a fixed uplink budget \(\beta\)~\eqref{eq:txfrac}, we ask whether replacing
modality-agnostic keep-ratios with contribution-aware keep-ratios improves
accuracy when the expected transmitted payload is held constant.
Every controlled comparison holds fixed the intra-modality compressor
\(\mathcal{C}\) and the training protocol, and changes only the mapping from
budget to per-modality keep-ratios \(\{r_m\}\).
Budget is accounted in expected kept smashed-activation floats,
\(\mathrm{TX}=\sum_m D_m r_m\), together with the matched cut-gradient support
(Section~\ref{sec:compression}).

\subsection{Baselines}
\label{sec:baselines}

To the best of our knowledge, no prior method allocates a shared payload budget
across modalities, whether in SL or otherwise: existing
communication-efficient methods compress each transmitted tensor independently
and leave the division of a shared budget among modalities unspecified. Our
central comparison is therefore between \emph{ModalShare} and the
modality-agnostic default it replaces, \emph{uniform} allocation, which assigns
every modality the same keep-ratio (\(r_m=\beta\) for all \(m\)) and represents the 
existing practice of compressing each modality independently. Both satisfy
the iso-budget constraint~\eqref{eq:isobudget} and differ only in how the shared
payload is split. Crucially, equal keep-ratios are not equal bandwidth: when the
smashed dimensions differ, uniform allocation already sends a fraction
\(q_m^{\mathrm{uni}}=D_m/\sum_k D_k\) of the payload to modality
\(m\)~\eqref{eq:q-unif}, so a modality with a larger cut silently receives a
larger share. ModalShare instead sets \(\{r_m\}\) from the server-side
contribution probe of Section~\ref{sec:ours}, under the same total budget.

\subsection{Datasets}
\label{sec:datasets}

\textbf{CREMA-D}~\cite{cao2014crema} is an audio-visual emotion recognition dataset 
containing synchronized speech and facial expression data. It comprises $7,442$ video 
clips spanning six emotion categories. The dataset is split into $6,698$ training 
samples and $744$ testing samples. 

\textbf{MVSA-Single}~\cite{niu2016sentiment} is an image-text dataset used 
for sentiment analysis. 
It includes a stratified subsample of
\(2{,}592\) pairs split \(60/20/20\) into \(1{,}555\) training, \(518\)
validation, and \(519\) held-out pairs, with the validation partition reported
as test accuracy.

\textbf{UCI~HAR}~\cite{anguita2013public} is a six-class human activity recognition
dataset from smartphone inertial sensors, collected from 30 volunteers. 
It comprises $10,299$ samples, split into $7,352$ training and $2,947$ testing samples.
Each sample provides two synchronized streams: accelerometer and gyroscope.

\subsection{Models}
\label{sec:models}

\noindent \textbf{Encoders.} All experiments use the multimodal SL setup of Section~\ref{sec:setup}: each modality has a client-side
encoder truncated at a fixed early cut, and the server completes the remaining
modality-specific layers, fusion, and classification.
Visual streams (CREMA-D video; MVSA image) use a ResNet-18 backbone.
MVSA text uses DistilBERT, fine-tuned end-to-end.
CREMA-D audio is encoded from the waveform as an STFT log-magnitude spectrogram
(\(22{,}050\)\,Hz, tiled to three seconds, \(n_{\mathrm{fft}}{=}512\), hop
length \(353\)), giving a single-channel \(257{\times}188\) input to a ResNet-18
encoder trained from scratch.
UCI~HAR uses a small 1D CNN per inertial stream.

\noindent \textbf{Fusion.} Fusion is concatenation followed by a linear classifier head, with no
cross-modal attention.
Unlike attention-based fusion, concatenation does not reweight streams after
the cut, so an accuracy difference between allocations at matched \(\beta\) is
attributable to the allocation itself.

\begin{table}[t]
\centering
\caption{Flattened cut dimensions at the early split and the payload share
induced by equal keep-ratios, \(q_m^{\mathrm{uni}}=D_m/\sum_k D_k\).}
\label{tab:cut_dims}
\setlength{\tabcolsep}{6pt}
\renewcommand{\arraystretch}{1.1}
\begin{tabular}{llrr}
\toprule
Dataset & Modality & \(D_m\) & \(q_m^{\mathrm{uni}}\) \\
\midrule
\multirow{2}{*}{CREMA-D} & video & \(200{,}704\) & \(50.7\%\) \\
                         & audio & \(195{,}520\) & \(49.3\%\) \\
\midrule
\multirow{2}{*}{MVSA}    & image & \(200{,}704\) & \(80.3\%\) \\
                         & text  & \(49{,}152\)  & \(19.7\%\) \\
\midrule
\multirow{2}{*}{UCI HAR} & accel & \(32\)        & \(50.0\%\) \\
                         & gyro  & \(32\)        & \(50.0\%\) \\
\bottomrule
\end{tabular}
\end{table}

\begin{table*}[t]
    \centering
    \caption{ModalShare against uniform under SplitFC, early-split
    multimodal SL. (a) Accuracy (\%) across compression rates,
    with the accuracy difference \(\Delta\); the shaded \(\Delta\) row is bold where
    the gain exceeds 1pp. (b) At \(\beta{=}0.2\), the modality the probe favors and
    the payload share \(q\) it receives (\(q^{\mathrm{mod}}\): ModalShare, \(q^{\mathrm{uni}}\): uniform), with the probe's contribution gap \(g\). The gain tracks the
    measured gap: large on CREMA-D and MVSA, near-zero on the balanced UCI~HAR.}
    \label{tab:results_main}
    
    \begin{subtable}{0.62\textwidth}
    \centering
    \setlength{\tabcolsep}{3.2pt}
    \renewcommand{\arraystretch}{1.15}
    \resizebox{\linewidth}{!}{%
    \begin{tabular}{ccrrrrrr}
    \toprule
    Dataset (no comp.) & Allocation & \multicolumn{6}{c}{Compression rate} \\
    \cmidrule(lr){3-8}
    & & \(2.5\times\) & \(3.3\times\) & \(5\times\) & \(10\times\) & \(20\times\) & \(40\times\) \\
    \midrule
    \multirow{3}{*}{CREMA-D ($53.1$)}
    & Uniform
    & \(46.9\){\scriptsize$\pm$6.2} & \(41.1\){\scriptsize$\pm$6.4}
    & \(32.2\){\scriptsize$\pm$5.2} & \(23.4\){\scriptsize$\pm$2.5}
    & \(19.4\){\scriptsize$\pm$1.6} & \(18.5\){\scriptsize$\pm$1.6} \\
    & ModalShare
    & \(55.2\){\scriptsize$\pm$2.1} & \(54.5\){\scriptsize$\pm$3.5}
    & \(47.6\){\scriptsize$\pm$5.9} & \(33.5\){\scriptsize$\pm$6.5}
    & \(25.0\){\scriptsize$\pm$4.8} & \(21.0\){\scriptsize$\pm$4.4} \\
    \rowcolor{gray!12}
    & \(\Delta\)
    & \(\mathbf{+8.3}\){\scriptsize$\pm$4.9}
    & \(\mathbf{+13.4}\){\scriptsize$\pm$4.1}
    & \(\mathbf{+15.4}\){\scriptsize$\pm$4.3}
    & \(\mathbf{+10.1}\){\scriptsize$\pm$5.1}
    & \(\mathbf{+5.6}\){\scriptsize$\pm$5.1}
    & \(\mathbf{+2.5}\){\scriptsize$\pm$4.9} \\
    \midrule
    \multirow{3}{*}{MVSA ($61.8$)}
    & Uniform
    & \(52.5\){\scriptsize$\pm$2.5} & \(49.1\){\scriptsize$\pm$3.8}
    & \(48.0\){\scriptsize$\pm$2.1} & \(42.4\){\scriptsize$\pm$3.3}
    & \(40.6\){\scriptsize$\pm$2.5} & \(40.2\){\scriptsize$\pm$2.6} \\
    & ModalShare
    & \(61.7\){\scriptsize$\pm$2.8} & \(60.1\){\scriptsize$\pm$4.6}
    & \(60.4\){\scriptsize$\pm$6.3} & \(51.0\){\scriptsize$\pm$7.4}
    & \(47.7\){\scriptsize$\pm$4.7} & \(41.0\){\scriptsize$\pm$3.3} \\
    \rowcolor{gray!12}
    & \(\Delta\)
    & \(\mathbf{+9.2}\){\scriptsize$\pm$2.5}
    & \(\mathbf{+11.0}\){\scriptsize$\pm$5.4}
    & \(\mathbf{+12.4}\){\scriptsize$\pm$6.4}
    & \(\mathbf{+8.6}\){\scriptsize$\pm$8.1}
    & \(\mathbf{+7.1}\){\scriptsize$\pm$5.5}
    & \(+0.8\){\scriptsize$\pm$4.5} \\
    \midrule
    \multirow{3}{*}{UCI HAR ($63.7$)}
    & Uniform
    & \(51.2\){\scriptsize$\pm$2.4} & \(47.3\){\scriptsize$\pm$2.6}
    & \(43.6\){\scriptsize$\pm$2.3} & \(39.4\){\scriptsize$\pm$2.1}
    & \(39.2\){\scriptsize$\pm$2.3} & \(37.3\){\scriptsize$\pm$2.2} \\
    & ModalShare
    & \(51.5\){\scriptsize$\pm$2.2} & \(48.0\){\scriptsize$\pm$2.5}
    & \(44.7\){\scriptsize$\pm$3.5} & \(40.2\){\scriptsize$\pm$3.1}
    & \(39.1\){\scriptsize$\pm$2.1} & \(37.5\){\scriptsize$\pm$2.1} \\
    \rowcolor{gray!12}
    & \(\Delta\)
    & \(+0.3\){\scriptsize$\pm$0.7}
    & \(+0.7\){\scriptsize$\pm$1.0}
    & \(\mathbf{+1.1}\){\scriptsize$\pm$1.3}
    & \(+0.8\){\scriptsize$\pm$1.7}
    & \(-0.1\){\scriptsize$\pm$1.6}
    & \(+0.2\){\scriptsize$\pm$0.6} \\
    \bottomrule
    \end{tabular}%
    }
    \subcaption{Accuracy versus uplink compression (no-compression in parentheses).}
    \label{tab:acc_vs_compression}
    \end{subtable}%
    \hfill
    \begin{subtable}{0.37\textwidth}
    \centering

    \setlength{\tabcolsep}{5pt}
    \renewcommand{\arraystretch}{1.2}
    \resizebox{\linewidth}{!}{%
    \begin{tabular}{cccccc}
    \toprule
    Dataset & Modalities & \(q^{\mathrm{mod}}\) & \(q^{\mathrm{uni}}\) & Gap \(g\) & \(\Delta\) \\
    \midrule
    \multirow{2}{*}{CREMA-D}
    & audio & \(0.99\) & \(0.49\) & \multirow{2}{*}{\(1.00\)} & \multirow{2}{*}{\(+15.4\)} \\
    & video & \(0.01\) & \(0.51\) & & \\
    \midrule
    \multirow{2}{*}{MVSA}
    & text  & \(0.88\) & \(0.20\) & \multirow{2}{*}{\(0.90\)} & \multirow{2}{*}{\(+12.4\)} \\
    & image & \(0.12\) & \(0.80\) & & \\
    \midrule
    \multirow{2}{*}{UCI~HAR}
    & accel & \(0.56\) & \(0.50\) & \multirow{2}{*}{\(0.12\)} & \multirow{2}{*}{\(+1.1\)} \\
    & gyro  & \(0.44\) & \(0.50\) & & \\
    \bottomrule
    \end{tabular}%
    }
    \subcaption{Probe contribution versus realised allocation.}
    \label{tab:alloc_calibration}
    \end{subtable}
    
    \end{table*}

\noindent \textbf{Splitting point.}
In realistic SL scenarios, edge clients cannot host most of the model
locally: computation relies mostly on the server, and the client only transmits
smashed activations from an early cut. 
That is also where the smashed-activation payload is large enough for an
uplink budget to bind and for per-modality keep-ratios to matter. 
Based on this, we decide to use the early split (layer-1) as the default, 
unless otherwise specified.
Concretely, the client holds only the first stage of each encoder: ResNet-18
stem plus the first residual block for CREMA-D video/audio and MVSA image;
DistilBERT layers \(0\)--\(1\) for MVSA text; the first 1D-CNN block for
UCI~HAR; and the server runs the remaining modality-specific layers, pooling,
fusion, and the classifier.
A late split instead keeps the full encoder on the client and exchanges only
pooled embeddings (\(256\)--\(512\) dimensions per modality), leaving little
payload to allocate.

Table~\ref{tab:cut_dims} lists the cut dimensions and the payload share that
equal keep-ratios induce for early split.
CREMA-D is nearly balanced; on MVSA the image stream already receives
four-fifths of the transmitted payload under the modality-agnostic default.

\subsection{Hyperparameters and evaluation}
\label{sec:hyperparameters}

CREMA-D is trained with SGD (learning rate \(10^{-3}\), weight decay
\(10^{-4}\), momentum \(0.9\)) for \(100\) epochs with a \(5\)-epoch
learning-rate warmup and a \(5\)-epoch compression ramp.
MVSA is trained with AdamW (learning rate \(3{\times}10^{-4}\), weight decay
\(10^{-3}\)) for \(30\) epochs with one-epoch warmups.
Batch size is \(64\) in both cases.
Allocation uses floor \(r_{\min}{=}0\), probe period
\(T_{\mathrm{probe}}{=}10\), EMA coefficient \(\alpha{=}0.3\), and temperature
\(\tau{=}1\), each shared across all datasets, budgets, and compressors.
Freezing uses confidence threshold \(\gamma{=}0.90\), streak length \(K{=}3\),
and earliest freeze epoch \(e_{\min}{=}10\), likewise shared.
For the underlying compressor, we use SplitFC~\cite{oh2025communication}, unless 
otherwise specified.
Final test accuracy is 
averaged over \(N{=}12\) seeds.


\noindent \textbf{Experimental Environment.} 
We implemented our framework and all baselines using the PyTorch library~\cite{paszke2019pytorch}.
All experiments were conducted on a server equipped with a 32-core AMD Ryzen Threadripper PRO CPU, 504 GB 
of memory and an NVIDIA GeForce RTX 4090 GPU.

\section{Evaluation Results}
\label{sec:results}

\subsection{Same budget, better accuracy}
\label{sec:results-main}

Table~\ref{tab:acc_vs_compression} reports accuracy against uplink compression
\(1/\beta\) under SplitFC with early split at matched total payload.
ModalShare strongly outperforms uniform on CREMA-D and MVSA at every
compression level.
UCI~HAR is milder but still favours the allocator at $5/6$ budgets: it settles at a 
slight keep-ratio imbalance rather than a corner, and that
mild reallocation yields a small accuracy gain over equal keep-ratios, suggesting
that the two sensor streams contribute roughly the same, so there is little
asymmetry to be exploited.

\begin{figure}[t]
\centering
\includegraphics[width=\linewidth]{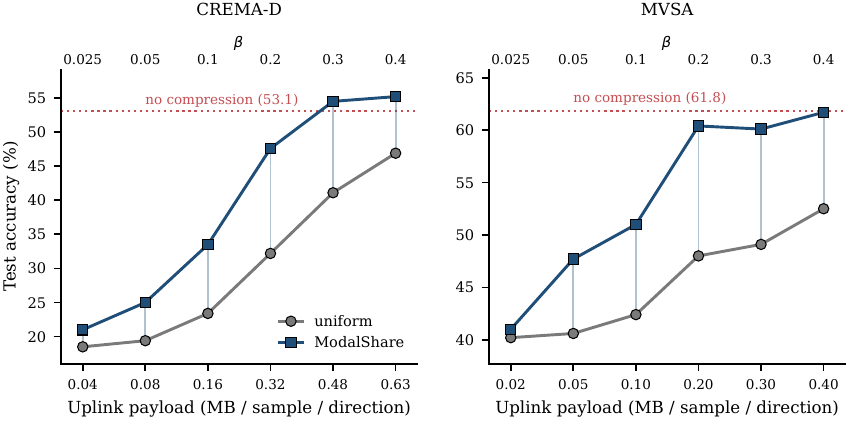}
\caption{Accuracy (\%) versus uplink payload.
Left: CREMA-D; right: MVSA. Red dotted: no-compression reference.
Top axis: bandwidth budget \(\beta\).}
\label{fig:acc_vs_payload}
\end{figure}

Figure~\ref{fig:acc_vs_payload} shows the same CREMA-D and MVSA comparison against
transmitted payload, so the horizontal axis is physical uplink cost rather than
a keep-ratio label.
Gains peak at moderate compression: the gap is largest near
\(5\times\) and mainly shrinks toward the tightest budgets 
(near the trivial floor).
Once the budget again carries usable signal, the same total payload is worth
more under ModalShare than under equal keep-ratios.

\noindent \textbf{Operating envelope.}
\label{sec:results-envelope}
Table~\ref{tab:acc_vs_compression} and Figure~\ref{fig:acc_vs_payload} show the
same pattern: \(\Delta\) peaks near moderate compression (about \(+15\)\,pp on
CREMA-D and \(+12\)\,pp on MVSA) and largely declines toward small budgets, approaching
near zero at \(40\times\).
At \(40\times\) both arms approach the trivial floor, so allocation has little
left to differentiate: CREMA-D sits just above six-class chance (\(16.7\)), and
MVSA uniform sits on the majority-class rate (\({\approx}40\)).
This pattern suggests that reallocation pays when the budget binds but still carries signal.

\subsection{Does the probe measure contribution?} 
\label{sec:results-calibration}
The gains above raise a question:
does ModalShare help \emph{because} it measures contribution, or would any
aggressive asymmetric split have helped on these datasets? On CREMA-D and MVSA
the two are observationally identical, so
Table~\ref{tab:alloc_calibration} places the probe's measurement beside the
allocation it produces. We summarise each dataset by the contribution gap
\(g\) of~\eqref{eq:confidence-gap}, i.e.\ the normalized separation of the
smoothed scores \(\{s_m\}\) that feed the keep-ratio map
(for \(M{=}2\), \(g=|s_A-s_B|/(s_A+s_B+\varepsilon)\)).
Thus \(g\) is a property of the probe scores \emph{before} they are mapped to
\(\{r_m\}\), not a restatement of the allocation. The allocation follows it: on
CREMA-D and MVSA the probe reads a large gap (\(g{\approx}1.0\) and \(0.90\)) and
concentrates the budget on the modality it scores as dominant (audio and text
respectively) for double-digit \(\Delta\); on the balanced UCI~HAR it reads a
small gap (\(g{\approx}0.12\)) and splits near-evenly. UCI~HAR is the
discriminating case: a method that forced asymmetry regardless of the data would
commit there too and would \emph{lose} accuracy, whereas ModalShare stays near
uniform because the probe finds little to reallocate. The allocation is thus a
function of the measured contribution, not a fixed asymmetric prior, which is
what ties the gains in Table~\ref{tab:acc_vs_compression} to a property the probe
reads in the data.

\subsection{Contribution, dimension skew \& modality selection}
\label{sec:results-decomposition}

On MVSA the image cut dimension is roughly four times the text cut dimension
(Table~\ref{tab:cut_dims}), so equal keep-ratios already place \(80.3\%\) of the
transmitted payload on image. This introduces a possible alternative reading of the MVSA
gain: that contribution plays no role and the allocator merely undoes a
dimension-induced skew. To separate the two effects we compare against a fixed
allocation that holds \(\mathrm{TX}\) constant but divides it differently. Where
uniform keep-ratios split the payload in proportion to each modality's cut
dimension, the \emph{equal-payload} control gives each modality half the
transmitted floats, isolating the dimension-skew component; notably, it is itself an
allocation that no existing compressor implements. ModalShare instead divides the
payload according to each modality's measured contribution.
Table~\ref{tab:mvsa_static_arms} reports the comparison at \(\beta{=}0.20\).

The gain decomposes into a skew component and a residual that no dimension-based rule can supply. Relative to uniform, moving to equal payload
recovers \(5.17\)\,pp of accuracy due to the dimension-skew component, and moving from
equal payload to ModalShare recovers a further \(7.23\)\,pp, for a total of
\(12.40\)\,pp. The larger share of the gain comes from the contribution component, a
part no fixed, dimension-based rule can supply, and which is precisely what 
ModalShare provides.

CREMA-D provides the complementary check (Table~\ref{tab:alloc_calibration}). 
Its two cut dimensions differ by about \(1.4\) percentage points in uniform
payload share (Table~\ref{tab:cut_dims}), so equal keep-ratios are already near
equal payload and the
dimension-skew component is absent by construction. Yet its contribution gain 
is the larger of the two datasets', which
establishes that the second component does not depend on unequal dimensions. The
dimension-skew term is thus specific to datasets with unequal cut dimensions,
while the contribution term is present in both.

A third reading remains: that the gain requires only identifying a winning modality, i.e. modality selection. Selection is in fact the $\tau \rightarrow 0$ limit of (\ref{eq:allocation-weights})-(\ref{eq:allocation-map}), so it is a point in the design space this formulation exposes rather than an alternative to it. Importantly, reaching that point still requires knowing which modality wins. Table V shows this is not readable at the cut: no per-stream signal recovers the allocation, and solo utility, which does see the fused model, moves the split by $0.02$ on CREMA-D against the probe's $0.50$. Selection also requires careful tuning, which the iso-budget map does not: the same allocation and freeze settings run unchanged across every dataset, budget, compressor, and split depth reported here, and on UCI HAR the map self-abstains at $g {\approx} 0.12$ where a forced commitment would cost accuracy.

\begin{table}[t]
\centering
\caption{MVSA static allocation references at \(\beta{=}0.20\).
\(q_{\mathrm{image}}/q_{\mathrm{text}}\) is the image/text float share.
\(\Delta\) is relative to uniform.
Equal payload (per modality) isolates dimension skew; the further gap to ModalShare
is the contribution effect.}
\label{tab:mvsa_static_arms}
\setlength{\tabcolsep}{5pt}
\renewcommand{\arraystretch}{1.12}
\begin{tabular}{lccr}
\toprule
Allocation & \(q_{\mathrm{image}}/q_{\mathrm{text}}\) & Acc.\ (\%) & \(\Delta\) (pp) \\
\midrule
Uniform          & \(0.803/0.197\) & 48.00{\scriptsize$\pm$2.1} & --- \\
Equal payload    & \(0.500/0.500\) & 53.17{\scriptsize$\pm$3.9} & +5.17 \\
ModalShare       & \(0.122/0.878\) & 60.40{\scriptsize$\pm$6.3} & +12.40 \\
\bottomrule
\end{tabular}
\end{table}

\subsection{Contribution proxies}
\label{sec:results-signals}

ModalShare's probe evaluates coalitions of modalities at the server, which costs
a forward pass. A natural question is whether a local signal, one
already available at the splitting point, can drive the same iso-budget
map~\eqref{eq:allocation-map} and recover the gain. Table~\ref{tab:proxies}
screens three at \(\beta{=}0.20\) under the early-SplitFC protocol used elsewhere:
the per-modality activation norm, the cut-gradient norm, and the solo utility
\(u(\{m\})\), which is the server's predictive utility when modality \(m\) is present and
the other are zeroed.

None moves the allocation to where it helps. On CREMA-D the strongest proxy,
activation-norm, shifts only part way toward audio
(\(q_{\mathrm{video}}{=}0.31\), against ModalShare's \(0.01\)) for a fraction of
the gain (\(+4.3\) vs \(+15.4\)\,pp); gradient-norm drifts the wrong way and solo
utility barely moves. 
On MVSA the proxy scores barely separate, so the allocation reaches only
near equal payload (\(q_{\mathrm{image}}{\in}[0.52,0.55]\)) and stays at or below uniform accuracy,
far from ModalShare's text corner (0.12, +12.4 pp).
UCI~HAR is near-null for all arms, as expected when there is
little to reallocate.

The failures have two distinct causes. The activation and gradient norms are
per-stream quantities computed before fusion; they cannot see how the modalities
combine in the prediction, so they do not track fusion-level contribution at all.
Solo utility does see the fused model, but measures each modality in
isolation. Its standalone predictive value is not what should set the
budget: a modality that is strong alone may be redundant once the other is
present, or weak alone yet complementary. Contribution is the \emph{marginal}
value a modality adds on top of the others, which only the coalition difference
in~\eqref{eq:shapley} captures. This is why ModalShare's full probe recovers the gain
where every cheaper signal falls short. 

\begin{table}[t]
  \centering
  \caption{Proxy screen at \(\beta{=}0.20\). For each dataset, \(\Delta\) is
  versus equal keep-ratios (pp) and \(q\) is the payload share of the superscripted
  modality; the uniform row is the \(\Delta{=}0\) reference.}
  \label{tab:proxies}
  \setlength{\tabcolsep}{3.8pt}
  \renewcommand{\arraystretch}{1.15}
  \begin{tabular}{lrcrcrc}
  \toprule
   & \multicolumn{2}{c}{CREMA-D} & \multicolumn{2}{c}{MVSA} & \multicolumn{2}{c}{UCI~HAR} \\
   \cmidrule(lr){2-3} \cmidrule(lr){4-5} \cmidrule(lr){6-7}
  Allocator
   & \(\Delta\) & \(q^{\mathrm{video}}\)
   & \(\Delta\) & \(q^{\mathrm{image}}\)
   & \(\Delta\) & \(q^{\mathrm{accel}}\) \\
  \midrule
  Shapley (ModalShare)
   & \(\mathbf{+15.4}\) & \(0.01\)
   & \(\mathbf{+12.4}\) & \(0.12\)
   & \(+1.1\) & \(0.56\) \\
  \midrule
  Activation-norm
   & \(+4.3\) & \(0.31\)
   & \(-1.5\) & \(0.55\)
   & \(+0.1\) & \(0.50\) \\
  Gradient-norm
   & \(-0.3\) & \(0.58\)
   & \(-1.3\) & \(0.52\)
   & \(0.0\) & \(0.51\) \\
  Solo utility
   & \(+0.3\) & \(0.50\)
   & \(-1.7\) & \(0.52\)
   & \(+0.1\) & \(0.50\) \\
  \midrule
  Uniform
   & \(0.0\) & \(0.51\)
   & \(0.0\) & \(0.80\)
   & \(0.0\) & \(0.50\) \\
  \bottomrule
  \end{tabular}
  \end{table}

\subsection{Allocation is orthogonal to the compressor}
\label{sec:results-coupling}

To test that the effect is not tied to one compressor, we re-run the identical
uniform-versus-ModalShare comparison under two well established compressors in SL: 
Top-\(S\)~\cite{yuan2020federated} and
RandTop-\(S\)~\cite{zheng2023reducing}, in addition to the adaptive 
SplitFC~\cite{oh2025communication} used elsewhere.
Allocation always selects the per-modality budgets; the compressor only
selects which coordinates survive inside each modality.
Table~\ref{tab:delta_by_compressor} reports the accuracy difference directly.
Across three compressors, three datasets, and four budgets, ModalShare
beats matched uniform in \(29\) of \(36\) configurations.
Most of the seven exceptions are within \(1\)\,pp on UCI~HAR, where contribution
is already near-balanced; the two large negatives are confined to CREMA-D at
mild compression under hard column selection.

\begin{table}[t]
\centering
\caption{Effect of ModalShare across intra-modality
compressors. Each cell is the accuracy difference
\(\Delta=\text{ModalShare}-\text{uniform}\) (percentage points).}
\label{tab:delta_by_compressor}
\setlength{\tabcolsep}{5pt}
\renewcommand{\arraystretch}{1.12}
\begin{tabular}{llrrrr}
\toprule
Compressor & Dataset & \multicolumn{4}{c}{Compression ratio} \\
\cmidrule(lr){3-6}
 & & \(5\times\) & \(10\times\) & \(20\times\) & \(40\times\) \\
\midrule
\multirow{3}{*}{SplitFC~\cite{oh2025communication}}
 & CREMA-D  & $+15.4$ & $+10.1$ & $+5.6$ & $+2.5$ \\
 & MVSA     & $+12.4$ & $+8.6$  & $+7.1$ & $+0.8$ \\
 & UCI HAR  & $+1.1$  & $+0.8$  & $-0.1$ & $+0.2$ \\
\midrule
\multirow{3}{*}{Top-$S$~\cite{yuan2020federated}}
 & CREMA-D  & $-5.11$ & $+7.71$  & $+7.67$ & $+4.97$ \\
 & MVSA     & $+6.52$ & $+13.21$ & $+9.91$ & $+7.26$ \\
 & UCI HAR  & $+0.08$ & $+1.32$  & $-0.21$ & $-0.04$ \\
\midrule
\multirow{3}{*}{RandTop-$S$~\cite{zheng2023reducing}}
 & CREMA-D  & $-2.51$ & $+4.99$  & $+9.65$  & $+8.18$ \\
 & MVSA     & $+4.21$ & $+11.50$ & $+10.05$ & $+8.93$ \\
 & UCI HAR  & $+0.77$ & $-0.40$  & $-0.05$  & $+0.03$ \\
\bottomrule
\end{tabular}
\end{table}

This failure mode is specific and identifiable: CREMA-D at \(5\times\) 
compression ratio, 
where Top-\(S\) loses \(5.11\)\,pp and
RandTop-\(S\) loses \(2.51\)\,pp relative to uniform.
At tighter budgets on the same dataset both hard sparsifiers \emph{gain}
(\(+4.97\) to \(+9.65\)\,pp).
Hard selection does not always leave a within-modality support that the
allocator can exploit at mild compression, and the effect is confined to the
dataset whose dimensions are already balanced. That is precisely where reallocation
has least to recover and most to disturb when the sparsifier is lenient.

Two further observations follow from the same table.
First, the effect is not an artifact of the adaptive compressor.
Under Top-\(S\) and RandTop-\(S\) on MVSA the gains are larger than under
SplitFC at several budgets, reaching \(13.21\)\,pp at \(10\times\), and two of
those Top-\(S\) configurations exceed the uncompressed reference, which could be 
attributed to the regularization effect of compression mechanisms, as already reported 
in~\cite{hinton2012improving,srivastava2014dropout, oh2025communication,zheng2023reducing}.
Second, the compressors themselves differ in strength: Top-\(S\) can
exceed SplitFC on absolute accuracy in some 
cases.
Compressor choice is an axis of its own, and every allocation comparison here is
within-compressor at matched \(\beta\).
We chose SplitFC as the primary underlying compressor because its uniform arm has the
highest absolute accuracy among the three compressors in most budgets.
Lastly and more importantly, this result highlights that existing SL compressors 
leave behind significant accuracy gains on the table in multimodal scenarios, 
with ModalShare being able to recover them.

\subsection{Dependence on split depth}
\label{sec:results-splitdepth}

\begin{figure}[t]
\centering
\includegraphics[width=\linewidth]{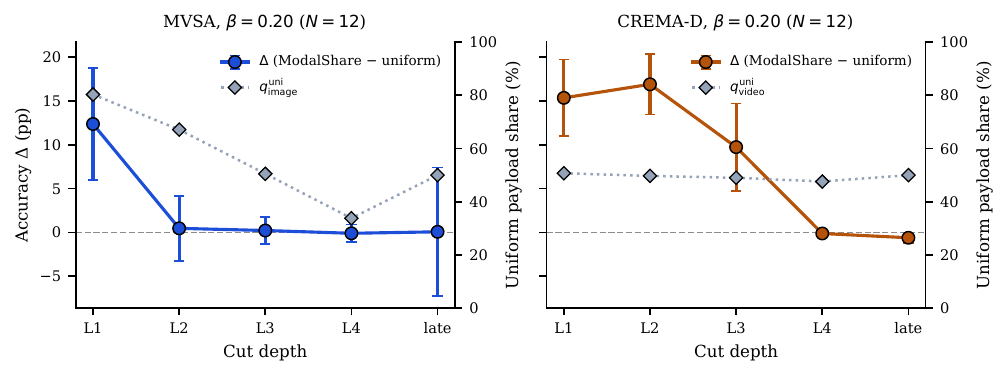}
\caption{Split-point ablation at \(\beta{=}0.20\).
Mean \(\pm\) std of \(\Delta=\text{ModalShare}-\text{uniform}\).
Left: MVSA; right: CREMA-D. Grey diamonds (right axis): uniform payload share of
modality~A.}
\label{fig:split_point}
\end{figure}

The early split is the natural operating point for split learning: it keeps the
client's share of the model small, which is the setting SL targets in the first
place, thus offloading most of the computation to the server. It
is also where the cut carries the most data, and therefore where dividing the
uplink across modalities is a decision at all. As the split moves server-ward the
client does more work and the smashed activations shrink toward a pooled
embedding, leaving less to allocate. We investigate the effect of varying the 
split depth at \(\beta{=}0.20\) and plot
\(\Delta\) against depth in Figure~\ref{fig:split_point}.

The two datasets behave differently with depth. On MVSA the gain is large at
layer\,1 and falls off quickly, and two effects fade together as it does: the
payload shrinks, and the dimension skew ModalShare corrects shrinks with
it. The uniform image share drops from \(80.3\%\) at layer\,1 to \(33.8\%\) at
layer\,4 (Figure~\ref{fig:split_point}, diamonds), so there is progressively less
skew to undo. CREMA-D, whose modality dimensions stay within \(3\%\) of equal at
every depth, has no such skew to begin with, yet still gains through
layer\,3. Because the uniform split is already near-balanced there, that gain
cannot be skew correction, therefore it is direct evidence that the effect is
contribution-driven, isolated on the dataset where dimension skew is absent by
construction.

Both datasets converge to zero \(\Delta\) at the late embedding cut, where almost
no payload crosses the boundary and there is nothing left to divide. Essentially, 
the gain
tracks the size of the decision it is making, and disappears as that decision
does. UCI~HAR shows little depth dependence throughout, consistent with its
near-balanced probe scores, and thus is not shown in the figure. 
Together these place ModalShare's main operating regime at the early
cut: the setting SL is built for, and the one where allocating the shared uplink
is an important decision.

\subsection{Freezing versus online}
\label{sec:results-freeze}

ModalShare's probe costs only server-side compute: every
\(T_{\mathrm{probe}}\) batches the server evaluates \(2^M\) coalitions (four
forward passes for \(M{=}2\)) to refresh \(\{r_m\}\). Once the allocation has
settled
on a stable split, those evaluations mostly reconfirm a decision already made.
This raises a practical question: can the probe be switched off once the split
is decided, and does doing so cost accuracy?

Table~\ref{tab:freeze_ablation} compares three arms at \(\beta{=}0.20\): 
uniform keep-ratios, an online ungated Shapley
diagnostic that probes throughout, and the confidence-freeze ModalShare used
in Table~\ref{tab:acc_vs_compression}.
Freezing locks \(\{r_m\}\) and stops probing once the contribution gap stays
decisive for \(K\) consecutive epochs after \(e_{\min}\)
(Section~\ref{sec:ours}), removing the residual server-side coalition cost.

\begin{table}[t]
\centering
\caption{Online versus early-freeze ModalShare at \(\beta{=}0.20\).
Mean \(\pm\) std of \(\Delta=\text{ModalShare}-\text{uniform}\).
The freeze arm adapts for a short post-ramp window, then locks \(\{r_m\}\) and
stops probing.}
\label{tab:freeze_ablation}
\setlength{\tabcolsep}{5pt}
\renewcommand{\arraystretch}{1.12}
\begin{tabular}{@{}lcccc@{}}
\toprule
Dataset & Uniform & Online & Freeze & \(\Delta_{\mathrm{f}-\mathrm{o}}\) \\
\midrule
CREMA-D
 & \(32.2{\pm}5.2\)
 & \(41.9{\pm}8.2\)
 & \(47.6{\pm}5.9\)
 & \(+5.7{\pm}6.2\) \\
MVSA
 & \(48.0{\pm}2.1\)
 & \(56.9{\pm}6.4\)
 & \(60.4{\pm}6.3\)
 & \(+3.5{\pm}6.9\) \\
\bottomrule
\end{tabular}
\end{table}

Confidence freeze does not give up the gain. On both datasets the freeze arm
matches or exceeds the online diagnostic in absolute accuracy while using the
same uniform reference as the main tables, and both beat uniform by a wide
margin.
CREMA-D separates early, so locking is cheap: the probe can stop once audio
dominates without waiting for the rest of training.
MVSA separates more slowly, which is why the deployed schedule waits for a
sustained gap rather than a fixed post-ramp clock. Freezing only after
evidence keeps the split from committing too early, while still cutting probe
cost (server overhead) for the remainder of training.

\section{Discussion}
\label{sec:discussion}

The formulation in Section~\ref{sec:method} is defined for arbitrary \(M\):
the Shapley probe evaluates coalitions over any modality set, and the
iso-budget map distributes payload across as many streams as
the client holds.
Our experiments cover \(M{=}2\), where a shared uplink already has to be
divided and the effect is cleanest to isolate.
Evidence is likewise strongest at early splits, where the cut carries enough
payload for allocation to be a decision at all; as the split moves
server-ward the gap closes
(Section~\ref{sec:results-splitdepth}), both because the uplink shrinks and
because the masking-based utility degrades once zeroing a modality produces an
off-distribution input.
The same link-local design places ModalShare under multi-client SL and Split
Federated Learning~\cite{thapa2022splitfed,hafi2024split}: each client runs its
own probe and keep-ratio map on its uplink, without changing how the server
aggregates client updates.
Cross-client contribution scoring and clients that hold different modality
subsets remain outside this paper.

We see multimodal contribution-aware allocation as an initial step towards a 
broader principle: a multimodal system should transmit only what the task
needs, both in centralized and collaborative settings.
Multimodal pipelines produce far more information than any uplink can carry,
and much of it is redundant once the other streams are present. This is what
a marginal-contribution score measures and a scalar compression rate cannot.
Scaling to \(M{\geq}3\) is the immediate step, where the allocation simplex has
interior structure that two streams cannot exhibit.

\section{Conclusion}
\label{sec:conclusion}
Multimodal SL inherits its compression from the unimodal case, where the only
decision available is how a single tensor is represented. This leaves a second
decision unmade: when several streams share one uplink, equal keep-ratios
divide the budget by cut dimension, a quantity unrelated to what each modality
contributes to the fused prediction.
Formulating that division as an iso-budget problem makes it a tunable degree of
freedom, and ModalShare shows that a contribution score measured at the server
is enough to set it well: \(15.4\) and \(12.4\)\,pp over equal keep-ratios at
matched payload, and near-uniform allocation where the measured gap is small.
The formulation admits any number of modalities and any fusion-level score.
Under a fixed link, how a multimodal system divides its budget is as
consequential as how tightly it compresses.

\clearpage
\bibliographystyle{IEEEtran}
\bibliography{references}

@article{oh2025communication,
  title={Communication-efficient split learning via adaptive feature-wise compression},
  author={Oh, Yongjeong and Lee, Jaeho and Brinton, Christopher G and Jeon, Yo-Seb},
  journal={IEEE Transactions on Neural Networks and Learning Systems},
  volume={36},
  number={6},
  pages={10844--10858},
  year={2025},
  publisher={IEEE}
}

@article{mudvari2024adaptive,
  title={Adaptive compression-aware split learning and inference for enhanced network efficiency},
  author={Mudvari, Akrit and Vainio, Antero and Ofeidis, Iason and Tarkoma, Sasu and Tassiulas, Leandros},
  journal={ACM Transactions on Internet Technology},
  volume={24},
  number={4},
  pages={1--26},
  year={2024},
  publisher={ACM New York, NY}
}

@article{xu2025contribution,
  title={Contribution-Guided Asymmetric Learning for Robust Multimodal Fusion under Imbalance and Noise},
  author={Xu, Zijing and Kou, Yunfeng and Wu, Kunming and Liu, Hong},
  journal={arXiv preprint arXiv:2510.26289},
  year={2025}
}

@inproceedings{wei2025improving,
  title={Improving multimodal learning via imbalanced learning},
  author={Wei, Shicai and Luo, Chunbo and Luo, Yang},
  booktitle={Proceedings of the IEEE/CVF International Conference on Computer Vision},
  pages={2250--2259},
  year={2025}
}

@inproceedings{zheng2023reducing,
  title={Reducing communication for split learning by randomized top-k sparsification},
  author={Zheng, Fei and Chen, Chaochao and Lyu, Lingjuan and Yao, Binhui},
  booktitle={Proceedings of the Thirty-Second International Joint Conference on Artificial Intelligence},
  pages={4665--4673},
  year={2023}
}

@article{yuan2020federated,
  title={A federated learning framework for healthcare iot devices},
  author={Yuan, Binhang and Ge, Song and Xing, Wenhui},
  journal={arXiv preprint arXiv:2005.05083},
  year={2020}
}

@article{vepakomma2018split,
  title={Split learning for health: Distributed deep learning without sharing raw patient data},
  author={Vepakomma, Praneeth and Gupta, Otkrist and Swedish, Tristan and Raskar, Ramesh},
  journal={arXiv preprint arXiv:1812.00564},
  year={2018}
}

@article{yuan2026communication,
  title={Communication-efficient multimodal federated learning: Joint modality and client selection},
  author={Yuan, Liangqi and Han, Dong-Jun and Wang, Su and Upadhyay, Devesh and Brinton, Christopher G},
  journal={IEEE Transactions on Mobile Computing},
  year={2026},
  publisher={IEEE}
}

@article{amini2025distributed,
  title={Distributed llms and multimodal large language models: A survey on advances, challenges, and future directions},
  author={Amini, Hadi and Mia, Md Jueal and Saadati, Yasaman and Imteaj, Ahmed and Nabavirazavi, Seyedsina and Thakker, Urmish and Hossain, Md Zarif and Fime, Awal Ahmed and Iyengar, SS},
  journal={arXiv preprint arXiv:2503.16585},
  year={2025}
}

@article{cheng2022greedy,
  title={Efficient modality selection in multimodal learning},
  author={He, Yifei and Cheng, Runxiang and Balasubramaniam, Gargi and Tsai, Yao-Hung Hubert and Zhao, Han},
  journal={Journal of Machine Learning Research},
  volume={25},
  number={47},
  pages={1--39},
  year={2024}
}

@article{lin2024split,
  title={Split learning in 6G edge networks},
  author={Lin, Zheng and Qu, Guanqiao and Chen, Xianhao and Huang, Kaibin},
  journal={IEEE Wireless Communications},
  volume={31},
  number={4},
  pages={170--176},
  year={2024},
  publisher={IEEE}
}

@inproceedings{parcalabescu2023mm,
  title={Mm-shap: A performance-agnostic metric for measuring multimodal contributions in vision and language models \& tasks},
  author={Parcalabescu, Letitia and Frank, Anette},
  booktitle={Proceedings of the 61st Annual Meeting of the Association for Computational Linguistics (Volume 1: Long Papers)},
  pages={4032--4059},
  year={2023}
}

@article{soenksen2022integrated,
  title={Integrated multimodal artificial intelligence framework for healthcare applications},
  author={Soenksen, Luis R and Ma, Yu and Zeng, Cynthia and Boussioux, Leonard and Villalobos Carballo, Kimberly and Na, Liangyuan and Wiberg, Holly M and Li, Michael L and Fuentes, Ignacio and Bertsimas, Dimitris},
  journal={NPJ digital medicine},
  volume={5},
  number={1},
  pages={149},
  year={2022},
  publisher={Nature Publishing Group UK London}
}

@article{hu2022shape,
  title={Shape: An unified approach to evaluate the contribution and cooperation of individual modalities},
  author={Hu, Pengbo and Li, Xingyu and Zhou, Yi},
  journal={arXiv preprint arXiv:2205.00302},
  year={2022}
}

@article{cao2014crema,
  title={Crema-d: Crowd-sourced emotional multimodal actors dataset},
  author={Cao, Houwei and Cooper, David G and Keutmann, Michael K and Gur, Ruben C and Nenkova, Ani and Verma, Ragini},
  journal={IEEE transactions on affective computing},
  volume={5},
  number={4},
  pages={377--390},
  year={2014},
  publisher={IEEE}
}

@inproceedings{niu2016sentiment,
  title={Sentiment analysis on multi-view social data},
  author={Niu, Teng and Zhu, Shiai and Pang, Lei and El Saddik, Abdulmotaleb},
  booktitle={International conference on multimedia modeling},
  pages={15--27},
  year={2016},
  organization={Springer}
}

@article{shapley1953value,
  title={A value for n-person games},
  author={Shapley, Lloyd S and others},
  year={1953},
  publisher={Princeton University Press Princeton}
}

@inproceedings{peng2022balanced,
  title={Balanced multimodal learning via on-the-fly gradient modulation},
  author={Peng, Xiaokang and Wei, Yake and Deng, Andong and Wang, Dong and Hu, Di},
  booktitle={Proceedings of the IEEE/CVF conference on computer vision and pattern recognition},
  pages={8238--8247},
  year={2022}
}

@inproceedings{wei2024enhancing,
  title={Enhancing multimodal cooperation via sample-level modality valuation},
  author={Wei, Yake and Feng, Ruoxuan and Wang, Zihe and Hu, Di},
  booktitle={Proceedings of the IEEE/CVF Conference on Computer Vision and Pattern Recognition},
  pages={27338--27347},
  year={2024}
}

@inproceedings{anguita2013public,
  title={A public domain dataset for human activity recognition using smartphones.},
  author={Anguita, Davide and Ghio, Alessandro and Oneto, Luca and Parra, Xavier and Reyes-Ortiz, Jorge Luis and others},
  booktitle={Esann},
  volume={3},
  number={1},
  pages={3--4},
  year={2013}
}

@article{paszke2019pytorch,
  title={Pytorch: An imperative style, high-performance deep learning library},
  author={Paszke, Adam and Gross, Sam and Massa, Francisco and Lerer, Adam and Bradbury, James and Chanan, Gregory and Killeen, Trevor and Lin, Zeming and Gimelshein, Natalia and Antiga, Luca and others},
  journal={Advances in neural information processing systems},
  volume={32},
  year={2019}
}

@inproceedings{li2025mbq,
  title={MBQ: Modality-Balanced Quantization for Large Vision-Language Models},
  author={Li, Shiyao and Hu, Yingchun and Ning, Xuefei and Liu, Xihui and Hong, Ke and Jia, Xiaotao and Li, Xiuhong and Yan, Yaqi and Ran, Pei and Dai, Guohao and others},
  booktitle={2025 IEEE/CVF Conference on Computer Vision and Pattern Recognition (CVPR)},
  pages={4167--4177},
  year={2025},
  organization={IEEE}
}

@article{hinton2012improving,
  title={Improving neural networks by preventing co-adaptation of feature detectors},
  author={Hinton, Geoffrey E and Srivastava, Nitish and Krizhevsky, Alex and Sutskever, Ilya and Salakhutdinov, Ruslan R},
  journal={arXiv preprint arXiv:1207.0580},
  year={2012}
}

@article{srivastava2014dropout,
  title={Dropout: a simple way to prevent neural networks from overfitting},
  author={Srivastava, Nitish and Hinton, Geoffrey and Krizhevsky, Alex and Sutskever, Ilya and Salakhutdinov, Ruslan},
  journal={The journal of machine learning research},
  volume={15},
  number={1},
  pages={1929--1958},
  year={2014},
  publisher={JMLR. org}
}

@article{meuwissen2026autoencoder,
  title={AutoEncoder-Compressed Parallel Split Learning for Pre-trained Model Fine-Tuning},
  author={Meuwissen, Bas and Tsouvalas, Vasileios and Meratnia, Nirvana},
  journal={arXiv preprint arXiv:2607.17913},
  year={2026}
}

@article{lin2026sl,
  title={Sl-acc: A communication-efficient split learning framework with adaptive channel-wise compression},
  author={Lin, Zehang and Lin, Zheng and Yang, Miao and Huang, Jianhao and Zhang, Yuxin and Fang, Zihan and Du, Xia and Chen, Zhe and Zhu, Shunzhi and Ni, Wei},
  journal={IEEE Transactions on Vehicular Technology},
  year={2026},
  publisher={IEEE}
}

@techreport{ericsson2026,
  title  = {Ericsson Mobility Report, June 2026},
  author = {{Ericsson}},
  institution = {Telefonaktiebolaget LM Ericsson},
  address = {Stockholm, Sweden},
  month  = {6},
  year   = {2026}
}

@inproceedings{mcmahan2017communication,
  title={Communication-efficient learning of deep networks from decentralized data},
  author={McMahan, Brendan and Moore, Eider and Ramage, Daniel and Hampson, Seth and y Arcas, Blaise Aguera},
  booktitle={Artificial intelligence and statistics},
  pages={1273--1282},
  year={2017},
  organization={Pmlr}
}

@inproceedings{thapa2022splitfed,
  title={{SplitFed}: When federated learning meets split learning},
  author={Thapa, Chandra and Arachchige, Pathum Chamikara Mahawaga and Camtepe, Seyit and Sun, Lichao},
  booktitle={Proceedings of the AAAI Conference on Artificial Intelligence},
  volume={36},
  number={8},
  pages={8485--8493},
  year={2022}
}

@article{hafi2024split,
  title={Split federated learning for 6G enabled-networks: Requirements, challenges, and future directions},
  author={Hafi, Houda and Brik, Bouziane and Frangoudis, Pantelis A and Ksentini, Adlen and Bagaa, Miloud},
  journal={IEEe Access},
  volume={12},
  pages={9890--9930},
  year={2024},
  publisher={IEEE}
}

@inproceedings{shao2020bottlenet++,
  title={Bottlenet++: An end-to-end approach for feature compression in device-edge co-inference systems},
  author={Shao, Jiawei and Zhang, Jun},
  booktitle={2020 IEEE International Conference on Communications Workshops (ICC Workshops)},
  pages={1--6},
  year={2020},
  organization={IEEE}
}

@inproceedings{hsieh2022c3,
  title={C3-SL: Circular convolution-based batch-wise compression for communication-efficient split learning},
  author={Hsieh, Cheng-Yen and Chuang, Yu-Chuan and Wu, An-Yeu},
  booktitle={2022 IEEE 32nd International Workshop on Machine Learning for Signal Processing (MLSP)},
  pages={1--6},
  year={2022},
  organization={IEEE}
}


\end{document}